\documentclass[runningheads]{llncs}
\usepackage[T1]{fontenc}
\usepackage{graphicx}
\usepackage{color}
\usepackage{hyperref} % added by PO
\usepackage{amsmath, amsfonts} %% added by PO

\usepackage[normalem]{ulem}
\useunder{\uline}{\ul}{}
\usepackage{multirow}
\begin{document}
\title{SiamTST: A Novel Representation Learning Framework for Enhanced Multivariate Time Series Forecasting applied to Telco Networks}
\titlerunning{SiamTST}
\author{
Simen Kristoffersen\inst{1} \and
Peter Skaar Nordby\inst{1} \and
Sara Malacarne\inst{2} \and \\
Massimiliano Ruocco\inst{1,3} \and
Pablo Ortiz\inst{2}
}
\authorrunning{S. Kristoffersen et al.}
\institute{Norwegian University of Science and Technology, Trondheim, Norway 
\and
Telenor Research, Fornebu, Norway\\
\email{\{sara.malacarne, pablo.ortiz\}@telenor.com}
\and
SINTEF Digital, Trondheim, Norway \\
\email{massimiliano.ruocco@sintef.no}\\
}
\maketitle              % typeset the header of the contribution
\begin{abstract}
We introduce \emph{SiamTST}, a novel representation learning fra\-me\-work for multivariate time series. SiamTST integrates a Siamese network with attention, channel-independent patching, and normalization techniques to achieve superior performance. Evaluated on a real-world industrial telecommunication dataset, SiamTST demonstrates sig\-ni\-fi\-cant improvements in forecasting accuracy over existing methods. Notably, a simple linear network also shows competitive performance, achieving the second-best results, just behind SiamTST.

The code is available at \url{https://github.com/simenkristoff/SiamTST}.

\end{abstract}
\section{Introduction}
\label{sec:introduction}

Multivariate Time Series (MTS) analysis keeps gaining importance in machine learning as industries generate increasingly extensive, complex datasets. These datasets feature diverse patterns and correlations, they vary immensely by industry, and present challenges due to the unlabelled nature of the data. Effective MTS analysis can yield significant insights, particularly in sectors like telecommunications, where understanding time series data is crucial for network traffic management and optimization.

The inherent complexity of MTS data, characterized by intricate temporal dynamics and spatial diversity, poses significant challenges in both global and local feature extraction. This complexity is exemplified in the telecommunication industry, where datasets include sensor readings across a network of cell towers in diverse geographic locations. These datasets provide a valuable opportunity for applying advanced machine learning techniques to address real-world problems.

Recent advances in representation learning have shown promise for enhancing MTS analysis. By employing pre-trained embeddings, these advances facilitate the extraction of complex patterns and improve performance in downstream tasks such as forecasting and classification. Inspired by successes in computer vision and natural language processing, this study introduces a novel framework tailored for telecommunication data, integrating contrastive learning methods and attention-based models.

We introduce SiamTST, a new deep learning architecture that combines attention mechanisms and Siamese networks \cite{9578004}, representing a significant novelty in the field. This architecture is designed to optimize performance in forecasting and transfer learning tasks across diverse telecommunication scenarios.

Key contributions of this research include:
\begin{itemize}
  \item \textbf{Novel Architecture}: Development of SiamTST, which integrates attention mechanisms and Siamese networks to enhance representation learning for MTS.
  \item \textbf{Implementation}: Public release of a PyTorch implementation of SiamTST, available in a well-documented code repository.
  \item \textbf{Comparative Analysis}: Extensive validation of SiamTST against state-of-the-art methods using a large-scale telecommunication dataset from Telenor.
  \item \textbf{Pre-training}: Investigation of the impact of pre-training on the model performance.
\end{itemize}

This study not only expands current knowledge in the field of MTS analysis but also offers practical insights that can be directly applied to improve operations and decision-making in the telecommunications industry.

\section{State of the Art}
\label{sec:sota}

Recently, Transformer models \cite{vaswani2017attention} have gained significant traction in the field of MTS analysis due to their capability to capture long-term dependencies and relationships across different parts of the input data. This section starts by discussing the pioneering TST \cite{zerveas2020transformerbased}, which has inspired notable subsequent works, including the iTransformer \cite{itransformer}, FedFormer \cite{fedformer}, Informer \cite{informer}, Pyraformer \cite{pyraformer} and the Autoformer \cite{autoformer}. We also highlight PatchTST \cite{patchtst}, which has been recognized for its strong performance in recent benchmarks.

The Transformer architecture has revolutionized sequence-to-sequence modeling by introducing full attention-based modeling.
The self-attention mechanism allows the model to dynamically weigh the importance of different elements in the sequence. The first significant adaption of the Transformer for MTS was the TST introduced in \cite{zerveas2020transformerbased}. The TST model modifies the original Transformer architecture to handle MTS data by removing the decoder component, only utilizing the Transformer encoder. The self-attention mechanism in TST allows the model to attend to all relevant parts of the sequence, capturing intricate patterns and dependencies between different features. TST is pre-trained unsupervised on masked time series modeling tasks. Here, random parts of the input sequence are masked, and the model is trained to reconstruct these masked values. This pre-training strategy enables the model to learn robust, fine-tuned representations for various downstream tasks such as classification and forecasting. However, TST incurs a high computational cost due to its self-attention mechanism, which has a complexity of $O(n^2)$, where $n$ is the length of the sequence being processed. This quadratic complexity means that longer context lengths significantly increase the computational burden, making training more resource-intensive and less efficient. Consequently, this high computational demand can limit the scalability of the model \cite{vaswani2017attention}.

PatchTST, a model proposed by \cite{patchtst}, tackles challenges in long time series with a novel patching approach. Inspired by ViT techniques \cite{visonTransformer}, PatchTST splits the time series into consecutive patches. This approach significantly reduces the sequence length and allows the model to capture local dependencies more effectively. PatchTST also introduces the concept of using channel independence, a concept successful in CNN and linear models, for Transformer models. In this approach, each variable in a multivariate time series is treated as an individual token, which is then combined in the final output layer. This method allows PatchTST to more effectively capture feature-specific patterns and interactions within the data \cite{patchtst}. In contrast, channel-mixing, where a single token integrates information from multiple variates, may hinder this detailed analysis. Through an ablation study across various MTS datasets, \cite{patchtst} demonstrates that processing each variate independently consistently outperforms channel-mixing in time series forecasting. However, a drawback of channel independence is that it may overlook potential interactions between different variables, which could be critical in capturing the full complexity of the MTS, useful for other downstreams tasks such as anomaly detection.

Contrastive learning methods for time series have shown impressive performance recently, with several papers pushing the SOTA forward. Contrastive methods learn representations by attracting similar instances together while pushing dissimilar instances away. This creates an embedding space that has proven helpful in improving performance across several downstream tasks, especially classification. One of the main challenges for time series representation learning is the retrieval of relevant samples. In \cite{franceschi2019unsupervised} they implemented an encoder architecture that introduced an unsupervised triplet loss. In this implementation, a reference sample was chosen, where the positive sample was a subseries of the reference, and the negative sample was a subseries of another random reference. This showed promising results, but ensuring that the positive sample is similar while also ensuring the negative sample is dissimilar is difficult across time series domains. In \cite{yue2021ts2vec} it was introduced TS2Vec, an effort to create universal representations for time series. TS2Vec hierarchically performs contrastive learning over augmented context views. This is achieved by sampling two overlapping subseries, which are then fed into an encoder where temporal contrastive loss and instance-wise contrastive loss are calculated. This sampling method is introduced to combat some of the challenges in earlier work where only subseries, temporal, or transformation consistency is considered. TS2Vec proposes contextual consistency as the solution, which treats the same timestamp in two augmented views as a positive pair. With its sampling technique, combined with a deep neural network encoder and dilated causal convolutions (DCCs) \cite{oord2016wavenet}, TS2Vec's performance delivered SOTA results at the time of its release, within both univariate and multivariate classification and forecasting. Due to its masking and contextual consistency, TS2Vec performed well despite missing input data. Contrastive learning has generally shown the best performance as a pre-processing step when classification is the chosen downstream task. According to \cite{zheng2023simts}, contrastive learning is inherently better at instance discrimination than future value prediction. This is supported by the fact that many representation learning methods for time series forecasting perform similarly to non-representation predictions \cite{hewamalage2022forecast}. SimTS, a framework from \cite{zheng2023simts}, aims to deliver better performance with forecasting as the designated downstream task. In their paper, they argue that negative pair sampling for time series is not generalizable across several different types of time series. SimTS does not use negative pair sampling at all, to avoid false repulsion. Instead, it focuses on maximizing the shared characteristics between positive samples, in this case, historical and future data. By introducing a Siamese network architecture and removing negative pairs, SimTS achieved SOTA performance for forecasting several public benchmarking datasets, including electricity transformer temperature \cite{informer}, which is the most common dataset for comparing methods. These results indicate that the representation learning framework needs to be carefully chosen based on the dataset's characteristics and the desired downstream task.
\section{Methods}
\label{sec:methods}

In this section, we explore the details of the proposed architecture to address our goal of creating valuable and robust representations of {MTS} data in the telecommunication sector.

\medskip
\textbf{Problem Definition.} We aim to enhance the performance of multivariate time series data analysis by developing improved representations compared to raw data. The problem can be formally defined as follows: given an input of a {MTS} in the form $X = [x_1, x_2, x_3, ..., x_L] \in \mathbb{R}^{N \times L}$, where $N$ denotes the number of variates and $L$ is the number of time steps, the objective is to create learned representations $Z = [z_1, z_2, z_3, ..., z_L] \in \mathbb{R}^{D \times L}$, with $D$ being a hyperparameter that defines the dimension of the learned representations, such that using $Z$ gives better results for a given task.

\medskip\textbf{Architectural Overview.} We address the problem of {MTS} representation learning by proposing a Siamese Time Series Transformer (\textbf{SiamTST}) inspired by PatchTST but with modifications to better suit pre-training. These modifications include the use of a Siamese pre-training architecture, {QKNorm} \cite{qk-norm} in the self-attention mechanism, removal of bias terms and pre-normalization in the Transformer encoder, randomized masking ratio, and replacing all normalization layers with {RMSNorm} \cite{rmsnorm}. The architecture of the model differs slightly between the pre-training and fine-tuning stages. We denote the part of the model that remains consistent throughout both stages as the \emph{backbone}. This backbone forms the foundation of SiamTST, ensuring that the learned representations can be adapted and fine-tuned for downstream tasks. This section will provide an overview of the architectures and their differences at both stages. In summary, the model operates as follows for each stage:

In the pre-training phase, the model aims to learn general representations from a collection of unlabeled time series. The model is trained using the masked time series modeling task. To enhance similar representations between identical time series, we utilize a Siamese architecture consisting of two identical models with shared parameters.

During the fine-tuning phase, the model is adjusted to a specific downstream task using labeled data. The fine-tuned version of SiamTST introduces task-specific output layers, referred to as heads, which are designed to refine the pre-trained representations for improved performance on the downstream task. All parameters from the pre-trained model are frozen, such that only the output head is updated.

\medskip\textbf{Channel-independence and Patching.}
The model takes a MTS as input, denoted by $X \in \mathbb{R}^{N \times L}$. Here $L$ is the length of the time series, and $N$ represents the number of variates. Inspired by the channel-independent patching mechanism in PatchTST, we split $X$ into $N$ univariate time series $x^{(i)} \in \mathbb{R}^{1 \times L}$, $i = 1,..., N$. Each time series $x^{(i)}$ is then split into patches. We achieve this by dividing $x^{(i)}$ into consecutive sub-series of length $P$, resulting in a sequence of non-overlapping patches denoted by $x_p^{(i)} \in \mathbb{R}^{P \times K}$. The value of $K$ is calculated as the integer division of $L$ and $P$, $\lfloor\frac{L}{P}\rfloor$, and represents the number of patches obtained after dividing the time series with no overlap \cite{patchtst}.

\begin{figure}[ht!]
    \centering
    \includegraphics[scale=0.3]{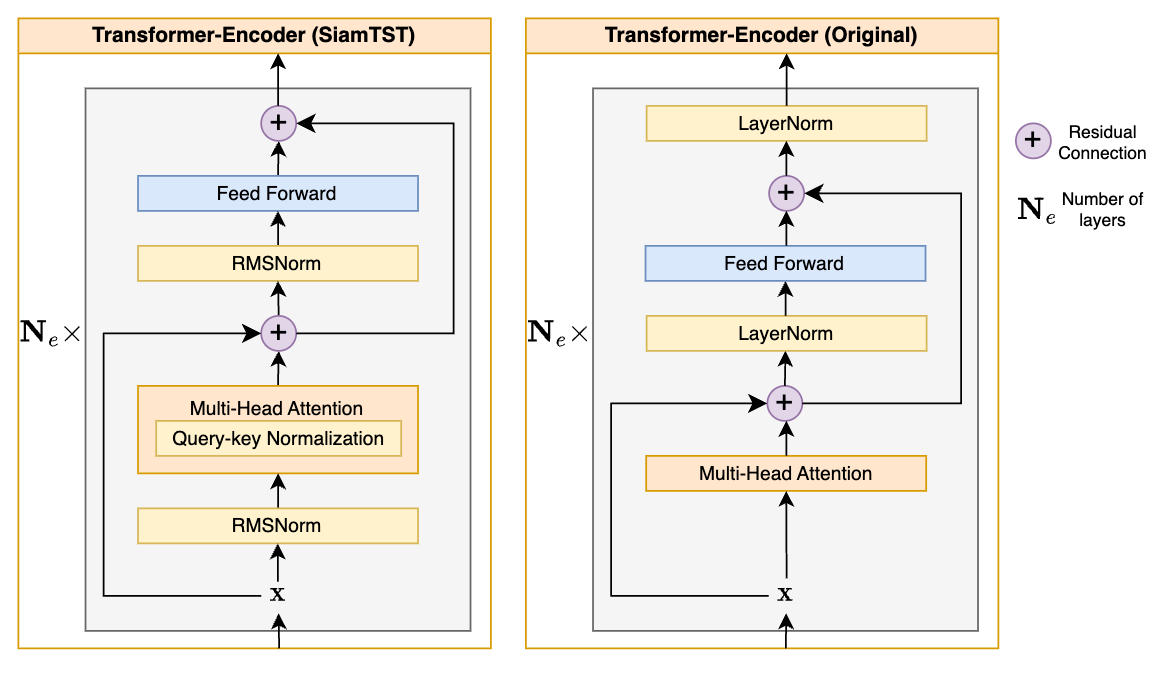}
    \caption{SiamTST's Transformer encoder module compared to the original Transformer encoder module from \cite{vaswani2017attention}. Our version moves the normalization layer ahead of the residual connections, replaces {LayerNorm} with {RMSNorm}, and applies {QKNorm} at the multi-head attention layer.}
    \label{fig:TransformerEncoder}
\end{figure}

\medskip\textbf{Backbone.}
At the core of our backbone is the Transformer encoder module illustrated in Figure \ref{fig:TransformerEncoder}. 

The module extends the original Transformer encoder \cite{vaswani2017attention}. However, we apply pre-normalization and replace all LayerNorm-layers with {RMSNorm}. Moving the normalization layers before the residual connection improves training stability because it allows the residual path to remain an identity map \cite{qk-norm}. Unlike LayerNorm, {RMSNorm} only involves re-scaling of the input and omits the re-centering property as this property contributes less to the model training. Thus, {RMSNorm} is comparable to LayerNorm in model performance while being computationally simpler and more efficient \cite{rmsnorm}.

\medskip
Before the patches are passed to the Transformer encoder module, each patch is projected into the latent dimension of the module, $D$. This is done by a trainable linear layer $W_p \in \mathbb{R}^{D \times P}$ with bias term $b_p \in \mathbb{R}^{D}$. To incorporate positional information, we add a matrix of learnable positional encodings, $W_{pos} \in \mathbb{R}^{D \times K}$, to the linear projection. The matrix is initialized with values sampled from a uniform distribution $\text{U}(0, 0.2)$, which are updated during training. These positional encodings inject information about the relative position of each time step within a patch \cite{zerveas2020transformerbased}. We denote the embedded patches as $x^{(i)}_d$, and the equation for the linear projection becomes:
\begin{equation}
x^{(i)}_d = W_px_p^{(i)} + b_p + W_{pos}, \quad x_d^{(i)} \in \mathbb{R}^{D \times K} \ .
\end{equation}

Note that the bias term $b_p$ is broadcasted across the matrix to match the dimensions of the linear projection output. This broadcasting approach is also applied to all subsequent bias terms throughout this paper.

The Transformer encoder module consists of $N_e$ encoder-layers, which processes $x_d^{(i)}$ consecutively. Each encoder layer is composed of two main layers: a multi-head attention mechanism and a {FFN}. Both layers are preceded by {RMSNorm} and succeeded by residual connections. These residual connections help maintain the gradient flow during training and ensure that the deeper layers receive meaningful gradients \cite{residualconnection}.

The multi-head attention layer applies self-attention to the input $x^{(i)}_d$.
Each attention head $h = \{1,...,H\}$ has a dimension of $d_k = \lfloor\frac{D}{H}\rfloor$. To obtain the query ($Q_h^{(i)}$), key ($K_h^{(i)}$) and value ($V_h^{(i)}$) pairs for each head, we use trainable linear layers $W_h^Q,W_h^K \in \mathbb{R}^{D \times d_k}$ and $W_h^V \in \mathbb{R}^{D \times D}$. In contrast to PatchTST's implementation, we do not include bias terms for these layers. We argue that bias terms are unnecessary as we already center the input by applying reversible instance normalization (RevIN) \cite{revin}, and thus they may introduce noise when pre-training on multiple {MTS}. Consequently, the following linear projections are applied:
\begin{align}
Q_h^{(i)} &= W_h^Q(x^{(i)}_d)^T, \quad Q_h^{(i)} \in \mathbb{R}^{D \times d_k}\ ,\\
K_h^{(i)} &= W_h^K(x^{(i)}_d)^T, \quad K_h^{(i)} \in \mathbb{R}^{D \times d_k}\ ,\\
V_h^{(i)} &= W_h^V(x^{(i)}_d)^T, \quad V_h^{(i)} \in \mathbb{R}^{D \times D}\ .
\end{align}

Next, scaled dot product attention (SDPA) is applied to the query, key, and value pairs to obtain the attention matrix, $A_h^{(i)}$. We use a modified {SDPA}, which incorporates {QKNorm}. {QKNorm}, as presented in \cite{qk-norm}, normalizes the queries and keys, making the softmax function in {SDPA} less prone to saturation. This avoids outputs close to the extreme values of $[0,1]$, which can inhibit the model's ability to learn diverse attention patterns. While \cite{qk-norm} proposes using the $L_2$-norm for normalization, we use {RMSNorm}. Thus, our attention function becomes:
\begin{gather}
\hat{Q}_h^{(i)} = RMSNorm(Q_h^{(i)})\ ,\\
\hat{K}_h^{(i)} = RMSNorm(K_h^{(i)})\ ,\\
(A_h^{(i)})^T = Attention(Q_h^{(i)}, K_h^{(i)}, V_h^{(i)}) = Softmax\left(\frac{\hat{Q}_h^{(i)}(\hat{K}_h^{(i)})^T}{\sqrt{d_k}}\right)V_h^{(i)}\ .
\end{gather}

We concatenate the attention from each head $A_h^{(i)}$ and obtain $A^{(i)} \in \mathbb{R}^{D \times D}$. The final attention is projected by a trainable linear layer $W_A \in \mathbb{R}^{D \times D}$ to integrate the attention from each head into a single representation. After multi-head attention, $x^{(i)}_d$ is updated by a residual connection $x^{(i)}_{d_2} = x^{(i)}_d + W_AA^{(i)}$, before being normalized and fed through the {FFN}.

The {FFN} in our model follows the standard Transformer's feed-forward structure. It consists of two trainable linear layers without bias term and a {GELU} activation in between as opposed to the usual {ReLU} activation. The {FFN} applies the following transformations to the input $x^{(i)}_{d_2}$:

\begin{align}
x^{(i)}_{ff_1} &= GELU(W_1x^{(i)}_{d_2})\ ,\\%, \quad x^{(i)}_{ff_1} \in \mathbb{R}^{D \times d_{ff}}\\
x^{(i)}_{ff_2} &= W_2x^{(i)}_{ff_1}\ ,%, \quad x^{(i)}_{ff_2} \in \mathbb{R}^{D \times d_{ff}}
\end{align}

where $W_1,W_2 \in \mathbb{R}^{D \times d_{ff}}$ represent the weights of the linear layers. The dimension of $d_{ff}$ is usually larger than $D$ to allow more expressive power and is typically set as a hyperparameter, in our case $d_{ff} = 4D$. After applying the {FFN}, we use another residual connection to update the input:

\begin{equation}
x^{(i)}_{d_3} = x^{(i)}_{d_2} + x^{(i)}_{ff_2}\ .
\end{equation}

After the {FFN} and residual connection, the output $x^{(i)}_{d_3}$ is passed to the next encoder layer in the Transformer encoder. This process is repeated for all $1,..,N_e$ encoder-layers to obtain the final encoded representations $z^{(i)} \in \mathbb{R}^{D \times K}$.
\section{Experimental setup}
\label{sec:experiments}

\subsection{Data}
\label{subsec:data}

The dataset is provided by Telenor Denmark and contains multiple multivariate time series with key performance indicators from cell towers across Denmark. These cell towers are responsible for routing and handling traffic for customers connected to the network using cellular devices. Each cell tower has three 120 degrees sectors that collect data separately. The data is aggregated to an hourly time resolution, and we used four months of data for this work. With 11661 unique sectors, each containing 13 features, this corresponds to approximately 458 million data points. 

\begingroup
\setlength{\tabcolsep}{10pt}
\begin{table}[ht]
\caption{Description of features included in the experiments.}
    \label{tab:Telenor_features}
    \centering
    \begin{tabular}{ll}
        Feature & Description \\
        \hline
        time\_period & Date and hour of record \\
        sector\_id & Unique ID of sector in cell tower \\
        mcdr\_denom & Total number of voice-related attempts \\
        msdr\_denom & Total number of data-related attempts \\
        {thp\_nom\_tt\_kpi} & {Nominator of the user throughput (data transferred)} \\
        thp\_denom\_tt\_kpi & {Denominator of the user throughput (duration)} \\
        {ho\_denom} & {Total number of handovers} \\
    \end{tabular}    
\end{table}
\endgroup
The selected features in the Telenor dataset are described in table \ref{tab:Telenor_features}. These features contain data regarding the origin of the time series data, as well as aggregated metrics from the given sector in a cell tower. The features describe voice- and data-related events, as well as throughput and handover. We disregarded other features because they have a much lower count, and therefore a much more digital-like and irregular behaviour, which makes them far less suitable for forecasting experiments.

\medskip%\noindent
\textbf{Pre-processing.} We processed our data in order to remove outlier sectors. First, we set a maximum threshold of 16 missing values for each sector, leaving us with $\sim$ 10200 unique sectors. We impute the missing values in the remaining sectors by interpolation.
To facilitate experimentation, we also chose a subset of sectors. We did so by dividing sectors in 100 clusters according to feature behaviour and selecting the sector closest to the centroid of each clusters. We checked that the resulting 100 sectors are representative for the geographical and data distributions. Last, early experiments showed two rather anomalous sectors that we removed, leaving us with 98 sectors. In all our experiments the splits follow a 60:20:20 proportion for train, validation and test sets, respectively.

\medskip%\noindent
\textbf{Normalization.} Network traffic volume exhibits significant variations across sectors, influenced by population density and usage patterns within specific geographic regions. Feature-wise normalization is applied for each sector. The transformation is fitted exclusively to the training set and then applied to the training, validation, and test sets, in order to avoid data leakage.

\subsection{Evaluation methods}
\label{subsec:evaluationMethods}

\textbf{LinearNet forecasting.} We implement the forecast head from PatchTST \cite{patchtst} as a barebone model and apply this model as a baseline reference, denoted \emph{LinearNet}. Here, each variate in a MTS is processed and forecasted as a univariate time series before each forecast is concatenated into a multivariate forecast.

LinearNet takes a multivariate time series $X \in \mathbb{R}^{N \times L}$ as input, where $N$ is the number of variates and $L$ is the number of time steps. The series is first instance normalized (RevIN), then $X$ is split into $N$ univariate time series $x^{(i)} \in \mathbb{R}^{1 \times L}$, $i = \{1,...,N\}$. Patching is applied, dividing the series into $K$ patches of size $P$, $x^{(i)}_p \in \mathbb{R}^{P \times K}$. Following the process of PatchTST, each patch is embedded into a $D$-dimensional vector through a linear layer with weights $W_p \in \mathbb{R}^{D \times P}$ and bias term $b_p \in \mathbb{R}^{D}$. Then, learnable positional encodings $W_{pos} \in \mathbb{R}^{D \times K}$ are added.

The embedded patches $x^{(i)}_d$ are flattened into a single vector $x^{(i)}_f \in \mathbb{R}^{1 \times (D \cdot K)}$ and passed to the final linear layer. This layer produces an univariate forecast sequence $\hat{y}^{(i)} = W_ox^{(i)}_f + b_o$ for $H$ time steps, where $W_o \in \mathbb{R}^{H \times (D \cdot K)}$ are the weights and $b_o \in \mathbb{R}^H$ is the bias term. Finally, the univariate forecasts $\hat{y}^{(i)}$ for all variates $i$ are concatenated into the multivariate forecast $\hat{Y} \in \mathbb{R}^{N \times H}$, and the instance normalization is reversed.

\medskip%\noindent
\textbf{Ridge regression forecasting.} Ridge regression is a linear regression model with an $L2$-regularization term of weight $\lambda$ \cite{hoerl1970ridge}. We forecast with ridge regression using the same method proposed in TS2Vec. We use a dataset of learned representations $X \in \mathbb{R}^{M \times L}$, with $L$ being the number of observations and $M$ being the feature space of the representations. With a forecast horizon $H$, we fit a ridge regression to forecast $\hat{Y} \in \mathbb{R}^{P \times H}$, where $P$ is the original feature space. We fit the model with the following values, $\lambda \in $ [0.1, 0.2, 0.5, 1, 2, 5, 10, 20, 50, 100, 200, 500, 1000], and select the best result based on the validation set \cite{yue2021ts2vec}.

\medskip%\noindent
\textbf{Metrics.} We use MAE and MSE to report forecasting performance.

\subsection{Experiments}
\label{sec:experimentalPlan}

\textbf{E1: Comparison with {SOTA}.} We choose a subset of {SOTA} models for {MTS} representation learning, namely TS2Vec \cite{yue2021ts2vec}, CoST \cite{woo2022cost}, SimTS \cite{zheng2023simts}, and PatchTST \cite{patchtst}, as well as the baseline models from Ridge Regression \cite{hoerl1970ridge} and LinearNet. We evaluate the performance of learnt representations by using them for the downstream task of forecasting. We use forecast horizons of 24, 48, 96, and 168 hours. For each forecast horizon, we train and evaluate the models 98 times, one time for each sector, and present the mean aggregated performance. 

It is important to note that TS2Vec, CoST, and SimTS differ from PatchTST and SiamTST in their forecasting methodologies. TS2Vec, CoST, and SimTS first extract learned representations and then fit these representations to a Ridge Regression model. In contrast, PatchTST and SiamTST integrate a linear layer on top of the pre-trained backbone that directly outputs the forecasted values.

The hyperparameters for all models are the default values from their respective public implementations as linked in their papers. For SiamTST we choose the same values as for PatchTST, except a fixed learning rate of 0.001 and a random masking ratio in the range $[0.15, 0.55]$, as opposed to 0.4 in PatchTST.

\medskip%\noindent
\textbf{E2: Pre-training.} In the second experiment we pre-train the backbone of SiamTST on multiple sectors (5, 10, 50, and 98). Then, we fine-tune the head of the pre-trained model and forecast 24, 48, 96, and 168 hours for each sector individually. We compare the performance to a baseline SiamTST model which is pre-trained and fine-tuned on each sector individually. The hypothesis is that including more sectors during the pre-training phase can improve the robustness and performance of the model's learned representation due to increased diversity and volume of training data.
The configuration is the same as for E1, except for increasing the number of epochs for the pre-training stage to 100, since the model is exposed to more observations during this phase. 
In this experiment, we present the mean aggregated score across all target sectors.
\section{Results and Discussion}
\label{cha:resultsAndDiscussion}

\subsection{Comparison with SOTA}
\label{sec:exp:SOTAComp}

Table \ref{tab:telenor_single_sector_results} displays the mean aggregated results from forecasting on the Telenor dataset. It can be seen that SiamTST consistently outperforms across all forecast horizons the other benchmarked methods TS2Vec, CoST, SimTS, PatchTST, and simpler baseline methods like LinearNet and Ridge Regression. The gap increases with longer forecast horizons, demonstrating SiamTST's superior capability in handling long-term dependencies. For example, at the 168-hour forecast horizon, SiamTST outperforms TS2Vec by $17.08\%$ in {MAE} and $22.99\%$ in {MSE}.
Interestingly, the second-best performer is our implementation of a simple neural forecaster, LinearNet. This model is identical to the forecasting head utilized by PatchTST and SiamTST for generating forecasts.

\begingroup
\setlength{\tabcolsep}{4.2pt}
\begin{table}[h]
\caption{Results from multivariate time series forecasting on the Telenor dataset, benchmarking several representation learning methods for forecast horizons of 24, 48, 96, and 168 hours. The best results are in \textbf{bold} and the second-best {\ul underlined}.}
\label{tab:telenor_single_sector_results}
\centering
\scalebox{1}{
\begin{tabular}{lccccc}
%\hline
Model & Metric & 24h & 48h  & 96h & 168h \\ \hline \\[-10pt]
\multirow{2}{*}{\textbf{SiamTST}} & MAE & \textbf{0.407}$\pm$0.093 & \textbf{0.415}$\pm$0.075 &  \textbf{0.426}$\pm$0.056 & \textbf{0.437}$\pm$0.040 \\
 & MSE & \textbf{0.544}$\pm$0.777 & \textbf{0.567}$\pm$0.587 & \textbf{0.573}$\pm$0.405 & \textbf{0.576}$\pm$0.270 \\[2pt]

\multirow{2}{*}{TS2Vec} & MAE & 0.495$\pm$0.120 & 0.507$\pm$0.121 & 0.520$\pm$0.123 & 0.527$\pm$0.128 \\
 & MSE & 0.717$\pm$0.605 & 0.745$\pm$0.623 & 0.752$\pm$0.604 & 0.748$\pm$0.581 \\[2pt]

\multirow{2}{*}{CoST} & MAE & 0.460$\pm$0.114 & 0.472$\pm$0.116 & 0.479$\pm$0.121 & 0.483$\pm$0.129 \\
  & MSE & 0.640$\pm$0.559 & 0.667$\pm$0.581 & 0.670$\pm$0.569 & 0.670$\pm$0.556 \\[2pt]

\multirow{2}{*}{SimTS} & MAE & 0.444$\pm$0.283 & 0.460$\pm$0.290 & 0.471$\pm$0.292 & 0.479$\pm$0.295 \\
 & MSE & 0.619$\pm$2.166 & 0.653$\pm$2.226 & 0.661$\pm$2.168 & 0.664$\pm$2.048 \\[2pt]

\multirow{2}{*}{PatchTST} & MAE & 0.459$\pm$0.094 & 0.454$\pm$0.074 & 0.468$\pm$0.054 & 0.484$\pm$0.037 \\
 & MSE & 0.620$\pm$0.779 & 0.627$\pm$0.586 & 0.639$\pm$0.402 & 0.651$\pm$0.265 \\[2pt]

\multirow{2}{*}{LinearNet} & MAE & {\ul 0.412}$\pm$0.094 & {\ul 0.423}$\pm$0.076 & {\ul 0.436}$\pm$0.057 & {\ul 0.446}$\pm$0.041 \\
 & MSE & {\ul 0.553}$\pm$0.769 & {\ul 0.580}$\pm$0.585 & {\ul 0.590}$\pm$0.405 & {\ul 0.595}$\pm$0.270 \\[2pt]

\multirow{2}{*}{Ridge} & MAE & 0.611$\pm$0.138 & 0.678$\pm$0.130 & 0.718$\pm$0.111 & 0.737$\pm$0.089 \\
  & MSE & 7.381$\pm$6.834 & 13.946$\pm$10.183 & 14.256$\pm$8.872 & 10.335$\pm$5.426
%\hline
\end{tabular}
}
\end{table}
\endgroup

\begin{figure}[!ht]
    \centering
    \includegraphics[scale=0.36]{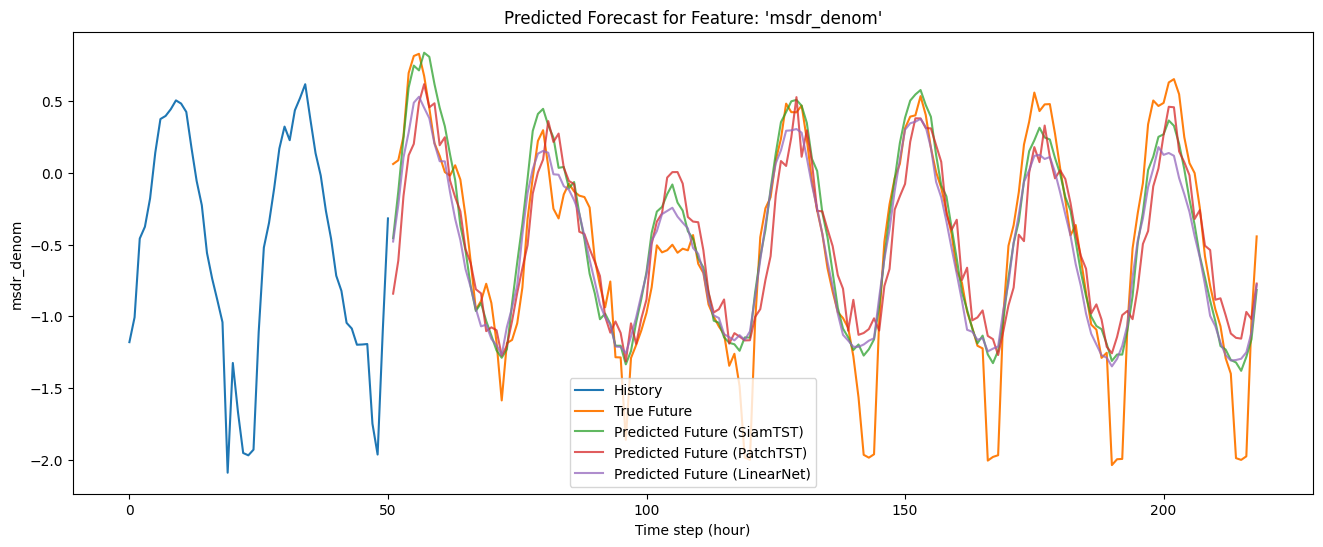}
    \caption{The figure shows 168-hour forecasts for the feature $msdr\_denom$ for a given sector. The forecasts are made by the following models: SiamTST, PatchTST, and LinearNet. The {orange line} displays the true future values.}
    \label{fig:experiment_1_forecasts_msdr_denom}
\end{figure}

Figure \ref{fig:experiment_1_forecasts_msdr_denom} provides a visual comparison of the forecasts from the top three performers, SiamTST, PatchTST and LinearNet, over a 168-hour horizon for the feature $msdr\_denom$. The true future values demonstrate the accuracy of SiamTST's predictions, particularly at the peaks.

The inclusion of LinearNet in the comparison provides a valuable perspective. LinearNet, despite its simplicity, outperforms several complex representation learning methods like TS2Vec, CoST, SimTS, and, most interestingly, PatchTST. The difference between PatchTST and LinearNet is how PatchTST uses an encoder to extract meaningful representations of time series data. Our finding questions the necessity of sophisticated representation learning frameworks as simpler models can achieve competitive performance. However, on the other hand, SiamTST's ability to surpass LinearNet confirms the value of advanced representation learning techniques, especially for capturing intricate patterns and long-term dependencies.

We note that the high values for the standard deviations in our experiments are mainly caused by the feature $ho\_denom$ (total number of handovers). Since the data is normalized according to the training set, when peak anomalies occur in the test set, they cause abnormally high errors. Despite this, we deliberately include the feature to challenge the representation learning schemes.
We validate the statistical significance of the performance differences by conducting a Student's t-test between SiamTST and the second-best performing model, LinearNet. The p-values confirm that the improvements in {MAE} and {MSE} are statistically significant when using a significance level of 0.05.

\subsection{Pre-training}
\label{sec:exp:preTraining}

Results from the pre-training experiment are presented in Table \ref{tab:telenor_multisector_pretraining_results}. We pre-trained our model backbone using different amounts of sectors and compared the results to a baseline, which has exclusively seen data from one sector during training. The results show an increase in model performance when more sectors are included in the training data. The best-performing model has 50 sectors included in the pre-training step, with the 98-sector model at a close second. The results have been validated with a t-test and all improvements are statistically significant with respect to the baseline.

\begingroup
\setlength{\tabcolsep}{5.4pt}
\begin{table}[h]
\caption{Results from multivariate time series forecasting on the Telenor dataset with a pre-trained backbone. The best results are  in \textbf{bold}, and the second-best {\ul underlined}.}
\label{tab:telenor_multisector_pretraining_results}
\centering
\scalebox{1}{
\begin{tabular}{lccccc}
Model & Metric & 24h & 48h  & 96h & 168h \\ \hline \\[-10pt]
\multirow{2}{*}{Baseline} & MAE & 0.407$\pm$0.093 & 0.415$\pm$0.075 & 0.426$\pm$0.056 & 0.437$\pm$0.040 \\
  & MSE & 0.544$\pm$0.777 & 0.567$\pm$0.587 & 0.573$\pm$0.405 & 0.576$\pm$0.270 \\[2pt]
  
\multirow{2}{*}{5 sectors} & MAE & 0.398$\pm$0.094 & 0.405$\pm$0.075 & 0.415$\pm$0.054 & 0.424$\pm$0.038 \\
  & MSE & 0.543$\pm$0.790 & 0.561$\pm$0.593 & 0.565$\pm$0.405 & 0.565$\pm$0.268 \\[2pt]

\multirow{2}{*}{10 sectors} & MAE & 0.393$\pm$0.094 & 0.400$\pm$0.075 & 0.410$\pm$0.055 & 0.419$\pm$0.040 \\
 & MSE & 0.538$\pm$0.792 & 0.555$\pm$0.595 & 0.559$\pm$0.408& 0.560$\pm$0.272 \\[2pt]

\multirow{2}{*}{50 sectors} & MAE & \textbf{0.388}$\pm$0.094 & \textbf{0.394}$\pm$0.074 & \textbf{0.403}$\pm$0.054 & \textbf{0.414}$\pm$0.038 \\
 & MSE & \textbf{0.529}$\pm$0.788 & \textbf{0.545}$\pm$0.591 & \textbf{0.547}$\pm$0.404 & {\ul 0.550}$\pm$0.268 \\[2pt]

\multirow{2}{*}{98 sectors} & MAE & {\ul 0.392}$\pm$0.093 & {\ul 0.397}$\pm$0.074 & {\ul 0.405}$\pm$0.053 & {\ul 0.415}$\pm$0.038 \\
 & MSE & {\ul 0.533}$\pm$0.789 & {\ul 0.547}$\pm$0.590 & {\ul 0.548}$\pm$0.402 & \textbf{0.549}$\pm$0.268

\end{tabular}
}
\end{table}
\endgroup
\vspace{-1cm}
\begin{figure}[h]
    \centering
    \includegraphics[scale=0.365]{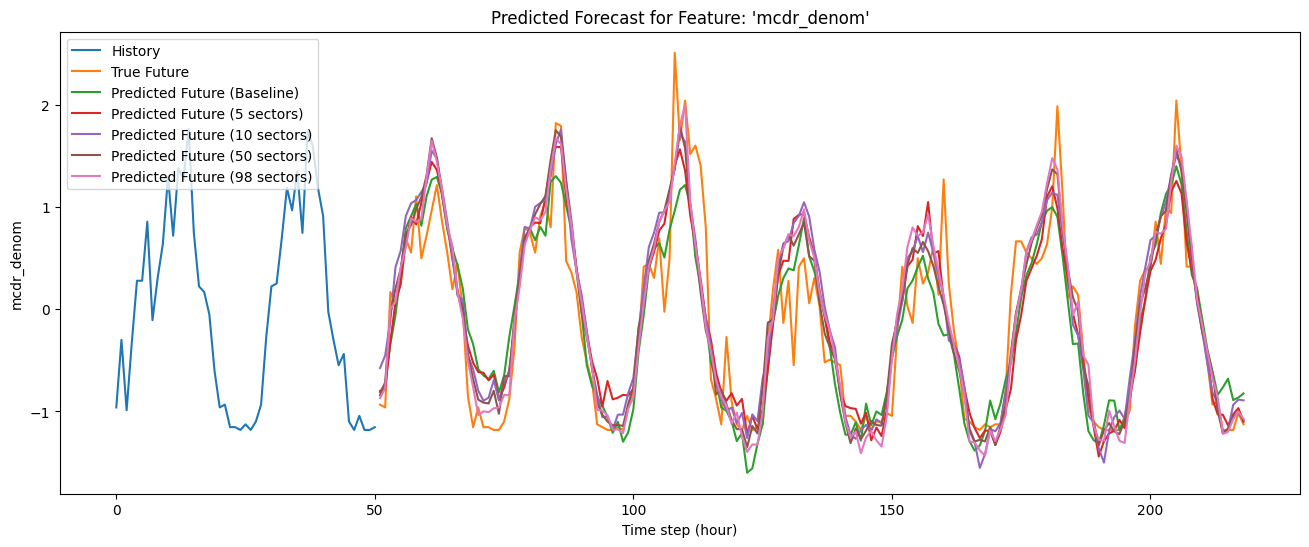}
    \caption{The figure shows 168-hour forecasts for the feature $mcdr\_denom$ for a given sector. The forecasts are made by the baseline model and models pre-trained on $5$, $10$, $50$, and $98$ sectors. The {orange line} displays the true future values.}
    \label{fig:experiment_2_forecasts_mcdr_denom}
\end{figure}

The results show that the learned representations perform better when more training data is available, but also that the improvement practically flattens after 50 sectors. We hypothesize that the model may have already reached an optimal point for the given data complexity and the number of model parameters.

When comparing the improvement of best model with respect to the baseline across forecasting horizons, {MAE} displays a stable increase of just over $5\%$ for all horizons. For {MSE}, the difference in performance increases with the forecasting horizon, with a relative improvement of $2.76\%$, $3,88\%$, $4.54\%$ and $4.69\%$ for horizons of 24, 48, 96, and 168 hours, respectively. This suggests that the model's robustness to errors is particularly improved for longer-term predictions when pre-trained on data from different sectors, as can be seen in Fig \ref{fig:experiment_2_forecasts_mcdr_denom}.
\section{Conclusion}
\label{sec:summary}
In this study we introduce SiamTST, a novel architecture for multivariate time series representation learning. This framework has been rigorously developed and evaluated, showcasing its ability to generalize across various cell towers and sectors in the telecommunication industry. SiamTST outperformed all other models, including those utilizing state-of-the-art contrastive learning and Transformers, across all measured metrics (MAE, MSE) and forecasting horizons (24, 48, 96, 168 hours). This superior performance validates SiamTST's effectiveness in leveraging advanced representation learning to enhance forecasting accuracy significantly. The pre-training experiments demonstrated the value of incorporating multiple sectors during the training phase. By increasing the diversity of training data, SiamTST's forecasting performance improved approximately 5\% over models trained on single sectors. This supports the hypothesis that broader pre-training leads to more robust and generalizable models. 
SiamTST establishes
new benchmarks for MTS representation learning within the telecommunication domain and suggests its applicability to broader MTS contexts. Further research could explore its adaptation and effectiveness across different industrial domains and other complex MTS analysis scenarios.

%%%%%%%%%%%%%%%%
\begin{credits}
\subsubsection{\ackname} We thank David Zsolt Biro for providing the Telenor Denmark dataset and for helpful discussions on the data. This work has been carried out as part of the master thesis by SK and PSN \cite{msc_thesis}. MR, SM and PO have been partially supported by the Research Council of Norway through the ML4ITS project\footnote{\url{https://ml4its.github.io/}} (312062).

\subsubsection{\discintname}
The authors have no competing interests to declare that are relevant to the content of this article.

\end{credits}
%%%%%%%%%%%%%%%%%

\bibliographystyle{splncs04}
\bibliography{SiamTST}

\end{document}